\documentclass[conference]{IEEEtran}
\IEEEoverridecommandlockouts
\usepackage{cite}
\usepackage{amsmath,amssymb,amsfonts}
\usepackage{algorithmic}
\usepackage{graphicx}
\usepackage{textcomp}
\usepackage{xcolor}
\usepackage[ruled,vlined]{algorithm2e}
\usepackage{booktabs} % 用于创建漂亮的表格线
\usepackage{multirow} % 用于合并单元格
\usepackage{caption}
\usepackage{amsthm}

\def\BibTeX{{\rm B\kern-.05em{\sc i\kern-.025em b}\kern-.08em
    T\kern-.1667em\lower.7ex\hbox{E}\kern-.125emX}}
\begin{document}

\title{Efficient CNN-LSTM based Parameter Estimation of L\'evy Driven Stochastic Differential Equations}

\author{\IEEEauthorblockN{Shuaiyu Li}
\IEEEauthorblockA{\textit{School of Information Science} \\
\textit{Kyushu University}\\
Fukuoka, 819-0395, Japan \\
li.shuaiyu.641@s.kyushu-u.ac.jp}
\and
\IEEEauthorblockN{Yang Ruan}
\IEEEauthorblockA{\textit{Graduate School of 
Bioresource and Bioenvironmental Sciences} \\
\textit{Kyushu University}\\
Fukuoka 819-0395, Japan \\
ahxcry9619@hotmail.com}
\and
\IEEEauthorblockN{Changzhou Long }
\IEEEauthorblockA{\textit{Graduate School of Science and Technology} \\
\textit{Univeristy of Tsukuba }\\
Ibaraki, 305-8577, Japan \\
ryu.choshu.su@alumni.tsukuba.ac.jp}
\and
\IEEEauthorblockN{Yuzhong Cheng}
\IEEEauthorblockA{\textit{Joint Graduate School of Mathematics for Innovation} \\
\textit{Kyushu University}\\
Fukuoka, 819-0395, Japan \\
cheng.yuzhong.129@s.kyushu-u.ac.jp}
\iffalse
\and
\IEEEauthorblockN{5\textsuperscript{th} Given Name Surname}
\IEEEauthorblockA{\textit{dept. name of organization (of Aff.)} \\
\textit{name of organization (of Aff.)}\\
City, Country \\
email address or ORCID}
\fi
}

\maketitle

\begin{abstract}
This study addresses the challenges in parameter estimation of stochastic differential equations (SDEs) driven by non-Gaussian noises, which are critical in understanding dynamic phenomena such as price fluctuations and the spread of infectious diseases. Previous research highlighted the potential of LSTM networks in estimating parameters of $\alpha$-stable Lévy driven SDEs, but faced limitations including high time complexity and constraints of the LSTM’s chaining property. To mitigate these issues, we introduce the PEnet, a novel CNN-LSTM based three-stage model that offers (a) an end-to-end approach with superior accuracy and adaptability to varying data structures, (b) enhanced inference speed for long-sequence observations through initial data feature condensation by CNN, and (c) high generalization capability, allowing its application to various complex SDE scenarios. Experiments on synthetic datasets confirm PEnet's significant advantage in estimating SDE parameters associated with noise characteristics, establishing it as a competitive method for SDE parameter estimation in the presence of Lévy noise.
\end{abstract}


\section{Introduction}
Stochastic Differential Equations (SDEs) have emerged as powerful tools for studying the dynamics of various phenomena influenced by inherent randomness in the real world. The application of SDEs spans multiple disciplines, including ecology~\cite{harris2013flexible}\cite{gloaguen2018stochastic}, finance~\cite{yilmaz2019stochastic}, and epidemiology~\cite{keeling2008methods}. For instance, The potential-based model \cite{wang2019continuous} in ecology  employs an SDE driven by Gaussian noise, with a drift term capturing the  Gaussian mixture attractor surface, to analyze and predict the movement patterns of wildlife. In all these types of systems, the dynamics are fully determined by the trend and noise-related parameters. Therefore, the inverse problem, estimating the SDE's parameters from observations, plays a key role in gaining insights into system trends, forecasting system dynamics, and facilitating informed decision-making processes, which has garnered considerable interest in recent years.

For SDEs driven by Gaussian noise, the closed-form likelihood functions enable Maximum Likelihood Estimation-based methods to effectively estimate the model parameters \cite{schneider2014maximum}\cite{delattre2013maximum}. However, in real-world phenomena, in addition to small-scale fluctuations, systems may also exhibit strong ‘jumps’ at random time points, as observed in stock prices and other similar processes. When modeling such phenomena, it is often more appropriate to consider non-Gaussian driven noises rather than Gaussian ones~\cite{iacus2007sde}. L\'evy processes, as a generalization of Gaussian processes, possess the flexibility to capture both small-scale fluctuations and large jumps at random times, making them powerful tools for characterizing complex stochastic phenomena. The jumps behavior, controlled by the parameterized L\'evy measure of the noise, leads to significantly different stochastic properties compared to the Gaussian case. As a result, it is crucial to develop methods that can offer accurate, reliable, and efficient parameter estimates in the Lévy case. However, while non-Gaussian noise has provided greater flexibility to SDE models, it has also introduced challenges in estimation. 

Unlike the Gaussian cases, L\'evy noises are generally not closed under convolution.  For example, the distribution of the Student-L\'evy noise is unknown unless the observation frequency $h=1$ (i.e., the time interval between observations), making it infeasible to use traditional Maximum Likelihood Estimation (MLE). In such cases, Quasi-likelihood estimation~\cite{masuda2019non}\cite{masuda2011quasi} is employed to estimate the system's parameter, utilizing the limiting behavior under high-frequency observations. However, this approach has a significant drawback of making strong assumptions about the observation frequency, namely, $h$ tending to zero. If this assumption cannot be satisfied, the estimated accuracy will be subject to significant errors.

In scenarios such as $\alpha$ stable L\'evy-driven Ornstein-Uhlenbeck (OU) processes, the drift parameter of the model can be estimated using least squares estimation (LSE)~\cite{hu2009least}\cite{long2009least}, which alone is insufficient for estimating all parameters of the system. A simultaneous estimation is achieved by the moment method through complex iterative algorithms~\cite{cheng2020generalized}, while it only has limited applicability and cannot be easily generalized to complex systems.
Furthermore, even in simple cases where MLE is applicable, the speed of the algorithms is typically slow due to the need for complex numerical integration in likelihood function calculations~\cite{fang2022end}.

In recent years, deep learning has attracted significant attention from researchers due to its powerful feature extraction capabilities, opening up new possibilities for SDE parameter estimation. For linear SDEs driven by Gaussian noises, a network structure based on multilayer perceptron (MLP) has shown good performance in estimating drift parameters and noise intensity. However, this method is limited by its simple network structure, which restricts its applicability to more complex SDE models. Additionally, even with increased input length and sample size, the performance of the method does not improve significantly~\cite{xie2007estimation}.

For more general driven noises, a two-step Parameter Estimating Neural Network (PENN) based on the vanilla long short-term memory (LSTM) neural network has shown promising results~\cite{wang2022neural} on $\alpha$-stable L\'evy-driven SDEs and demonstrates excellent performance in processing sequences of lengths up to 500. However, when the length of input data increases, the sequential nature of LSTM poses considerable challenges, resulting in a substantial decrease in prediction speed. Additionally, the memory requirements for training become increasingly burdensome and difficult to accommodate.
\iffalse
Moreover, the performance of PENN on SDEs driven by more complicated noises which are not closed under convolution (e.g., Student-L\'evy noise) still requires further empirical verification.
\fi
For SDE models that are flexible enough, the parameter estimation often requires input sequences of sufficient length. Taking the Switching SDE as an example, where the system exhibits multiple hidden states and the transitions between states are governed by a continuous-time Markov process, a sufficiently long observation length is necessary to ensure the exploration of each hidden state, thus obtaining a full parameter estimation \cite{kohs2022markov}. Currently, there is a lack of parametric estimation method that can simultaneously estimate all parameters from variable-length long sequences for general L\'evy driven SDEs.

To address this issue, this paper proposes a three-stage parametric estimation approach based on the convolutional neural network (CNN) and LSTM, named PEnet, which offers the following advantages:
\begin{enumerate}
    \item End-to-end capability: The method does not rely on handcrafted features~\cite{gentili2021characterization}\cite{wagner2017classification} or preprocessing steps~\cite{argun2021classification}, providing a seamless and automated parameter estimation process.
    \item Flexibility in handling variable-length sequences: The approach can accommodate varying lengths of input sequences and observation intervals, which is particularly advantageous in scenarios where the underlying driven noise is not closed under convolution, surpassing the limitations of traditional methods.
    \item Efficiency: The method is capable of processing long sequences efficiently, thereby enabling parameter estimation for more complex systems.
    \item Generalizability: By modifying the input data and labels during training, the proposed network structure can be applied to parameter estimation tasks of other SDE models, demonstrating its generalizability.
\end{enumerate}
The paper is organized as follows. In Section 2, an overview of L\'evy-driven SDEs and several representative driven noises is provided, along with a detailed description of the proposed network structure. Section 3 presents the numerical experiments conducted to validate the accuracy and scalability of our proposed method. Section 4 concludes the paper by summarizing our findings and discussing future works.

\section{Methodology}
\iffalse
In this section, we first specify the considered SDEs driven by three representative noises. Then the proposed networks will be detailed.
\fi
\subsection{L\'evy Driven SDEs}
A stochastic differential equation is a mathematical equation that describes the evolution of a stochastic process. It is an extension of ordinary differential equations (ODEs) that incorporates random disturbances. The general form of a parameterized SDE is given by:
\begin{equation}
    dX(t)=a_{\mathbf{\Theta}_a}(X(t),t)dt + b_{\mathbf{\Theta}_b}(X(t),t)dW_{\mathbf{\Theta}_{Noise}}(t),
\end{equation}
where $X(t)$ represents the system state at time $t$, $a_{\mathbf{\Theta}_a}(X(t),t)$ and $b_{\mathbf{\Theta}_b}(X(t),t)$ are deterministic functions with parameters $\mathbf{\Theta}_a$ and $\mathbf{\Theta}_b$ which are vectors capturing the system trend and diffusion respectively. $dW_{\mathbf{\Theta}_{Noise}}(t)$ is a random term determined by parameter $\mathbf{\Theta}_{Noise}$ and can be regarded as the infinitesimal increment of driving noise at time t, representing the random disturbance on the system. 

The dynamics of the system is totally determined by the packaged parameter
\begin{equation}
    \mathbf{\Theta}:=[\mathbf{\Theta}_a^T,\mathbf{\Theta}_b^T,\mathbf{\Theta}_{Noise}^T]^T=[\theta_1,\ldots,\theta_M]^T,
\end{equation}
where $M$ is the number of considered parameters and the superscript $T$ is the transpose
operator. We aim to estimate the parameter $\mathbf{\Theta}$ of the system from discrete observed data $\mathbf{x}$. The sample trajectory $\mathbf{x}=[x_{t_1},\ldots,x_{t_N}]^T$ is observed at equally spaced discrete time points $\{t_j\}_{j=1}^N$. We denote the data frequency by $h$, and thus the observation time points and terminal time are $t_j=jh$ and $T=Nh$ respectively.

\subsection{Examples of Driven Noises}
Lévy noises or Lévy processes more precisely, are a class of stochastic processes that play a fundamental role in the theory of stochastic analysis and its applications. A Lévy process is a continuous-time process with stationary and independent increments, signifying that the magnitudes of changes within any given time interval are independent of the changes observed in other non-overlapping time intervals ~\cite{ken1999Levy}. The subsequent section provides an overview of several prominent L\'evy driven noises, on which the efficacy of the proposed estimation method will be validated. 
\begin{itemize}
    \item Wiener Process
\end{itemize}
The most well-known and widely studied Lévy process is the Wiener process, also known as the Brownian motion or Gaussian noise. A Wiener Process $W(t)$ is a L\'evy process with Gaussian distributed increments, namely for $t,u>0$,
\begin{equation}
    W(t+u)-W(t)\sim \mathcal{N}(0,u).
\end{equation}

The Wiener process employed here does not incorporate learnable parameters; instead, it serves as a limiting case of other noise sources for comparative purposes.

\begin{itemize}
    \item $\alpha$-stable L\'evy noise
\end{itemize}
The $\alpha$ stable Lévy process is a generalization of Brownian motion and encompasses a wide range of stochastic processes with heavy-tailed and asymmetric distributions. The $\alpha$-stable Lévy process is characterized by stable probability distributions, which exhibit stable scaling properties under addition. This means that the sum of independent $\alpha$-stable random variables remains $\alpha$-stable.
Here we consider a standard $\alpha$-stable L\'evy motion $Z_{\alpha}(t)$~\cite{janicki1997approximation}\cite{applebaum2009levy}, where the increments are stable distributed, namely for $0\leq s\leq t$,
\begin{equation}
    Z_{\alpha}(t)-Z_{\alpha}(s)\sim S_\alpha\left((t-s)^{\frac{1}{\alpha}}\right).
\end{equation}

A random variable $X$ is stable distributed, denoted as $X\sim S_\alpha(\epsilon)$, if it has characteristic function 
\begin{equation}
    \phi_X(u)=\exp\{-\epsilon^\alpha|u|^\alpha\},
\end{equation}
where $\alpha$ and $\epsilon$ are the stability index and noise intensity respectively. The jumping behavior of $X$ is controlled by $\alpha$ and it reduces to Gaussian with variance $2\epsilon^2$ when $\alpha=2$. Note that if $X\sim S_\alpha(1)$, then $\epsilon X\sim S_\alpha(\epsilon)$, through which the noise intensity parameter influences the amplitude.
\begin{itemize}
    \item Student L\'evy noise
\end{itemize}
For both the Wiener process and the $\alpha$-stable process, their distributions are closed under convolution. This means that their increments $\Delta=X(t+h)-X(t)$ belong to the same distribution family as the stochastic process itself at any given observation frequency $h$. However, for more general L\'evy noise, this property does not always hold, namely the increments of the random term may not necessarily follow the same distribution as the random term itself and may even be unknown, which poses significant challenges to traditional parameter estimation methods based on the distribution function. As we aim to develop parameter estimation methods for sufficiently flexible SDE models, we inevitably have to deal with such situations. 

In this study, the Student L\'evy noise is chosen as a representative of the case for its high representativeness ~\cite{massing2019stochastic} and the challenges associated with its parameter estimation ~\cite{massing2018simulation}. 
\iffalse
Several key reasons support this choice:
\begin{enumerate}
    \item The traditional parameter estimation methods for Student L\'evy driven SDEs becomes highly challenging when the distribution is unknown for observation frequencies where $h\neq 1$~\cite{massing2018simulation}. The good performance of the proposed method in this scenario holds significant significance.
    \item As a member of the GH family, the Student L\'evy process exhibits close connections with several important processes. For example, as the degrees of freedom increase, the Student L\'evy process tends to approach a Gaussian process, while approaching a Cauchy process as the observation frequency $h\rightarrow 0$~\cite{massing2019stochastic}. The experimental results obtained on the Student L\'evy process under different parameter settings are highly representative.
\end{enumerate}
\fi

Here we consider Student L\'evy driven noise $J(t)$, where $J(1)\sim t(\nu)$. A random variable $X$ is Student-t distributed, denoted as $X\sim t(\nu)$, if it has density function
\begin{equation}
        f(x;\nu)=\frac{\Gamma(\frac{\nu+1}{2})}{\Gamma(\frac{\nu}{2})\sigma^2\sqrt{\pi\nu}}\left(1+\frac{x^2}{\nu}\right)^{-\frac{\nu+1}{2}}.
\end{equation}
The parameter $\nu>0$, referred to as the degrees of freedom, determines the kurtosis and heavy-tailedness of the distribution. The effectiveness of the proposed method on the Student L\'evy case will be validated with a comparison with a traditional Cauchy quasi-maximum likelihood estimation (CQMLE) through subsequent numerical experiments.

\subsection{The network structure}

This study proposes a three-stage parameter estimation approach, known as PEnet, which simultaneously estimates all parameters $\Theta$ of the target SDE given discrete observations $\mathbf{x}$ and its observation frequency $h$: 
\begin{equation}
   \hat{\Theta}=PEnet(\mathbf{x},h). 
\end{equation}

\begin{figure*}[t]
  \centering
  \includegraphics[width=0.6\textwidth]{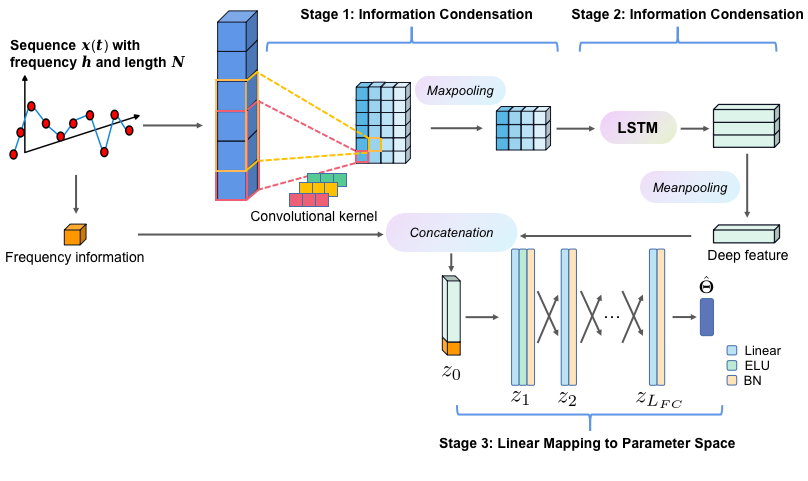}
  \caption{The architecture of the proposed method}
  \label{fig:1}
\end{figure*}

The architecture of PEnet is illustrated in Fig. \ref{fig:1}. Initially, a 1D CNN is utilized to condense the information from the raw long input sequences \cite{tang2020rethinking}. This type of CNN, specifically designed for processing sequential data, consists of convolutional and pooling layers \cite{li2021survey}. The convolutional layers employ learnable kernels to map the original 1D input to a high-dimensional feature space, generating a series of feature maps through window operations and convolution computations. 

Following the convolutional layers, max pooling layers are applied to reduce the feature maps' length, retaining essential features while condensing information, decreasing the original input length from $n$ to a much smaller $n'$. This not only lowers the time complexity for the next stage but also alleviates the issues encountered by LSTM neural network when handling long sequences.

In the second stage, the condensed information obtained from the first stage undergoes deep feature extraction using a LSTM neural network \cite{hochreiter1997long, greff2016lstm, yu2019review}. The LSTM, a prevalent tool in natural language processing, is adept at extracting deep features from variable-length time series. This stage utilizes $L$ LSTM cells to recursively translate the output from the preceding layer into hidden states or deep features for the next layer.
\iffalse
The intricate workings of the $l$-th LSTM layer are depicted in Fig. \ref{fig:lstm1} and Fig. \ref{fig:lstm2}, with further calculations detailed in \cite{wang2022neural}.

\begin{figure}[t]
  \centering
  \begin{minipage}[b]{0.2\textwidth}
    \centering
    \includegraphics[width=\textwidth]{lstm1.png}
    \caption{The stacked LSTM of the second stage.}
    \label{fig:lstm1}
  \end{minipage}
  \hfill
  \begin{minipage}[b]{0.27\textwidth}
    \centering
    \includegraphics[width=\textwidth]{lstm2.png}
    \caption{The LSTM cell where the superscripts indicating layer IDs are omitted.}
    \label{fig:lstm2}
  \end{minipage}
\end{figure}
\fi
The LSTM extracts hidden features forming a matrix of dimensions ${D_{lstm}} \times n'$, which are then condensed using a mean pooling operator to capture global information, resulting in a vector $\mathbf{\bar{y}}_{lstm}$ of length $D_{lstm}$. 
\iffalse
The mathematical representation is as follows:
\begin{equation}
    \mathbf{\bar{y}}_{lstm} = \frac{1}{n'}\sum_{i = 1}^{n'}\mathbf{y}_i,
\end{equation}
\fi
To enhance the model's accuracy in parameter estimation of the underlying SDE, the observation frequency information, denoted as $h$, is integrated at this stage. This information is pivotal as varying observation frequencies can correspond to different SDE families even with identical observed data. The final output feature of this stage is obtained by concatenating $h$ with the vector extracted by the LSTM, as expressed below:
\begin{equation}
     \mathbf{z_0} = [\mathbf{\bar{y}}_{lstm}^T,h]^T.
\end{equation}

 In the third stage, vector $\mathbf{z_0}$ is then passed through a fully connected neural network with $L_{FC}+1$ layers, mapping the extracted deep feature $\mathbf{z_0}$ to the parameter space. For the each layer of this stage, the input $z_{k-1}$ and output $z_{k}$ can be described as follows:
\begin{equation}
    \mathbf{z}_k=\begin{cases}
    f_{ELU}(f_{BN}(f_{LN}^{(k)}(\mathbf{z}_{k-1}))), &k=1,\\
    f_{ELU}(f_{LN}^{(k)}(\mathbf{z}_{k-1})), &k=2,\ldots,L_{FC},\\
    f_{LN}^{(k)}(\mathbf{z}_{k-1}), &k=L_{FC}+1,\\
    \end{cases}
\end{equation}

where $f_{LN},f_{ELU}$ and $f_{BN}$ are the linear (LN) layer, the activation function of exponential linear units (ELU) \cite{clevert2015fast} and the batch normalization (BN) layer \cite{ioffe2015batch}, respectively. The ELU function is a smoothed version of the ReLU function~\cite{nair2010rectified}, which performs better than ReLU and tanh activation functions on
capturing nonlinear mappings\cite{clevert2015fast}. The BN layer standardizes input features in every training batch, which has the effect of stabilizing the training procedure and accelerating the rate of loss reduction. This allows for the utilization of a relatively large learning rate, which in turn speeds up the training process~\cite{wang2022neural}. To prevent the network from becoming excessively flexible and impeding convergence, batch normalization is only applied to the first block of the network. 

We adopt a weighted $L1$ loss for network training. 
\begin{equation}
    L(\hat{\mathbf{\Theta}},\mathbf{\Theta})=\sum_{i=1}^M\lambda_i|\hat{\theta_i}-\theta_i|.
\end{equation}
$\{\theta_i\}$ and $\{\hat{\theta}_i\}$ are the $M$ components of $\mathbf{\Theta}$ and $\hat{\mathbf{\Theta}}$ respectively. The weights $\{\lambda_i\}$ are used to balance the scale differences among different parameters which are determined in advance by the training range, avoiding slow convergence of certain components. Moreover, $L1$ loss is more robust to outliers compared to $L2$ loss, making it suitable for training data with multiple sources of noise. 
\section{Numerical Experiments}

In this section, we evaluate the performance of PEnet in three different scenarios to demonstrate its effectiveness. Specifically, we compare the performance of PEnet with other machine learning-based estimation methods (i.e., PENN and MLP) in the Gaussian and $\alpha$ stable noise scenarios, and also with the traditional CQMLE in the case of Student L\'evy noise.

The experimental results demonstrate that PEnet exhibits a superior ability to capture essential information related to jumps in the data, resulting in more precise and robust parameter estimation of the stochastic component. Importantly, PEnet achieves exceptional computational efficiency, significantly reducing the training time per epoch to an average of 164.3 seconds on an NVIDIA GeForce RTX 3080 12GB GPU for a dataset containing 200,000 samples. In contrast, PENN requires an average training time of 640.8 seconds per epoch, and MLP, while faster with an average training time of 20.2 seconds per epoch, significantly lags behind both PEnet and PENN in terms of performance.
\iffalse
These compelling results underscore PEnet's competitiveness in tackling parameter estimation tasks, showcasing its exceptional accuracy and efficiency compared to other methods. PEnet emerges as a highly promising and practical solution for addressing challenges in modeling complex stochastic systems effectively.
\fi

\subsection{Experimental configuration}

To generate each sample in the experiment dataset, we uniformly select parameter values $\mathbf{\Theta}$, sample size N and time span T from the predetermined training ranges. Subsequently, we employ the Euler-Maruyama method~\cite{kloeden1992stochastic} to generate sample paths under these selected parameter values, which are then added to the dataset. In the $\alpha$ stable experiment, the $\alpha$-stable L\'evy noise is generated using the Chambers-Mallows-Stuck method~\cite{weron1996chambers}\cite{chambers1976method}, while the modified rejection method \cite{devroye1981computer} is employed to generate the Student L\'evy noise.

For all experiments, a consistent network architecture was employed, and the network configurations are outlined in table \ref{networkconfig}. The networks were implemented using the PyTorch library \cite{paszke2019pytorch}, and optimization was carried out using the ADAM method \cite{kingma2014adam} with a learning rate of 0.001.
\begin{table}[b]
    \centering
    \caption{Network Configuration}
    \begin{tabular}{l|l}
        \hline
        \textbf{Parameter} & \textbf{Value} \\
        \hline
        Number of convolutional layer $L_{cov}$ & 2 \\
        Size of convolutional kernels $k_{cov}$ & 3$\times$1 \\
        Number of convolutional kernels $D_{cov}$ & 25  \\
        Number of LSTM layers $L_{LSTM}$ & 4 \\
        Length of LSTM hidden state $F_{LSTM}$ & 25  \\
        Number of fully connected layers $L_{FC}$ & 3 \\
        Number of neurons in linear layers $L_{FC}$ & 20\\
        \hline
    \end{tabular}
    \label{networkconfig}
\end{table}

\subsection{Case 1: Gaussian case}
\label{Gaussian exp}

In this section, we assess the effectiveness of PEnet on Gaussian-driven OU processes. Since Gaussian noise does not involve any learnable parameters, this section primarily highlights PEnet's capability to capture the system's trend parameters and the noise intensity. We consider the following SDE with parameters listed in table \ref{table:params}:
\begin{align}
\label{gaussian}
dX(t) = -\eta X(t)dt &+ \epsilon dW(t), \ \ \  t\in[0,+\infty).\\
W(t)&\sim N(0,t)
\end{align}

\begin{table}[t]
    \centering
    \caption{Parameter of Gaussian driven SDE}
    \begin{tabular}{l|l|l|l}
        \hline
        \textbf{Parameter} & \textbf{Range} & \textbf{Parameter} & \textbf{Range} \\
        \hline
        Spanning Time $T$ & [5, 15] & Length $N$ & [3000, 4000]\\
        Drift $\eta$ & [0, 5] & Noise Intensity $\epsilon$ & [0, 0.05]\\
        \hline
    \end{tabular}
    \label{table:params}
\end{table}

 The model is determined by parameters $\mathbf{\Theta} = \{\eta,\epsilon\}$, and therefore the output of the network is two-dimensional vector $\hat{\mathbf{\Theta}}=[\hat{\eta},\hat{\epsilon}]^T$. To train the network, $K_{train}=200000$ samples are generated from table \ref{table:params}, and three paths with different lengths and time spanning are shown in Fig. \ref{fig:Gaussian path}. 

\begin{figure}[t]
  \centering
  \includegraphics[width=0.35\textwidth]{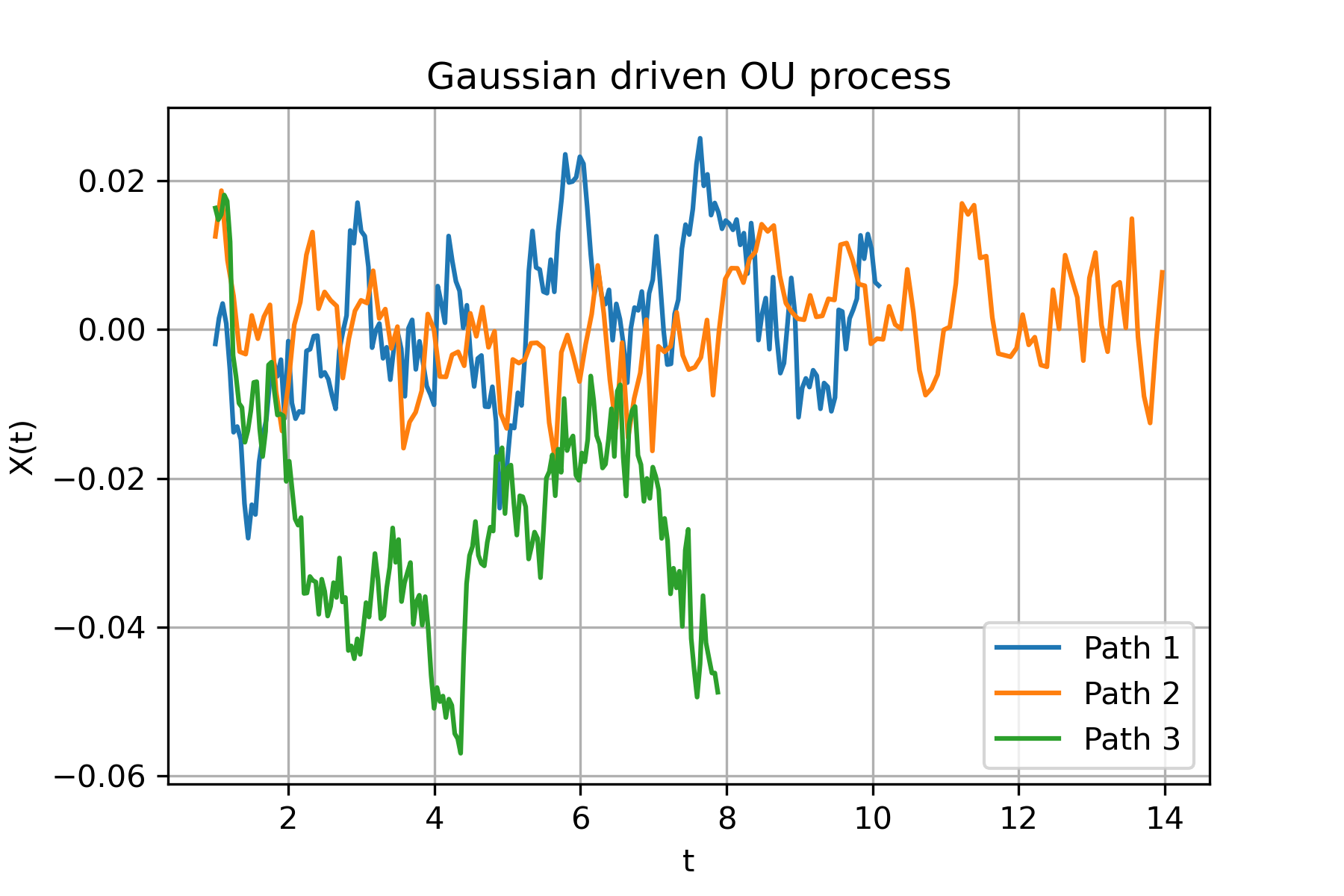}
  \caption{Three paths of Gaussian-driven OU processes}
  \label{fig:Gaussian path}
\end{figure}

\begin{figure}[t]
  \centering
  \includegraphics[width=0.4\textwidth]{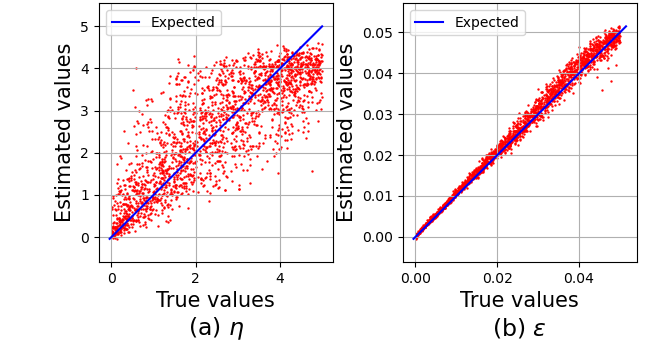}
  \caption{$\hat{\eta}$ and $\hat{\epsilon}$ estimated by the PEnet on 200,000 trajectories of OU process driven by Gaussian noise.}
  \label{fig:Gaussian predict}
\end{figure}

\iffalse
\begin{figure}[t]
  \centering
  \includegraphics[width=0.2\textwidth]{Gaussianplot.png}
  \caption{Three paths of Gaussian-driven OU processes}
  \label{fig:Gaussian path}
\end{figure}
\begin{figure}[t]
  \centering
  \includegraphics[width=0.2\textwidth]{gaussian.png}
  \caption{The estimated drift parameter $\eta$ and noise intensity $\epsilon$ by the PEnet on 200,000 trajectories of OU process driven by Gaussian noise.}
  \label{fig:Gaussian predict}
\end{figure}
\fi

The estimation performance of the trained network is tested on the test set of size $K_{test} = 5000$ and summarized by Fig. \ref{fig:Gaussian predict}. One can see that when the true value of the system trend $\eta$ falls within the middle region of the range $[0,5]$,  it can be estimated unbiasedly, but with a relatively larger deviation compared with the case when the true value approaches the boundary of the range. On the other hand, the estimation is biased close to the boundary of the training range. Specifically, when $\eta$ approaches 5, the estimated values tend to underestimate, while it tend to be overestimated when $\eta$ approaches 0. This discrepancy is attributed to the lack of extreme cases in the training set. On the other hand, when estimating the noise intensity parameter $\epsilon$, PEnet yields good unbiased estimation across the entire range [0, 0.05] of true values.

\begin{table*}[htbp]
\centering
\caption{Estimation errors of PEnet, PENN and MLP-based method on Gaussian driven OU process}
\label{error1}
\begin{tabular}{|c|c|c|c|c|c|c|}
\hline
\textbf{Method} & \textbf{$\eta$} & \multicolumn{2}{c|}{\textbf{$\hat{\eta}$}} & \textbf{$\epsilon$} & \multicolumn{2}{c|}{\textbf{$\hat{\epsilon}$}$\times 0.01$} \\
  & & \textbf{Mean $\pm$ SD} &\textbf{MAE} &  &\textbf{Mean $\pm$ SD}  &\textbf{MAE} \\
\hline
\textbf{PEnet} & 1.5 & $\mathbf{1.832}$±0.816 & 0.661 & 0.003 & 3.101±$\mathbf{0.166}$ & $\mathbf{0.156}$ \\

\textbf{PENN} & 1.5 & 1.912±\underline{0.616} & \underline{0.551} & 0.003 & 3.354±0.186 & 0.355 \\

\textbf{MLP} & 1.5 & 1.874±0.806 & 0.657 & 0.003 & \underline{3.058}±0.187 & 0.419 \\
\hline
\textbf{PEnet} & 2.5 & $\mathbf{2.692}$±0.809 & 0.695 & 0.001 & 1.072±$\mathbf{0.047}$ & $\mathbf{0.077}$\\

\textbf{PENN} & 2.5 & 2.801±\underline{0.677} & \underline{0.595} & 0.001 & 1.143±0.071 & 0.146 \\

\textbf{MLP} & 2.5 & 2.779±0.889 & 0.765 & 0.001 & \underline{0.989}±0.175 & 0.138 \\
\hline
\textbf{PEnet} & 3.5 & 3.343±0.678 & 0.549 & 0.003 & $\mathbf{3.121}$±$\mathbf{0.160}$ & $\mathbf{0.162}$ \\

\textbf{PENN} & 3.5 & \underline{3.527}±\underline{0.566} & \underline{0.467} & 0.003 & 3.144±0.163 & 0.176 \\

\textbf{MLP} & 3.5 & 3.427±0.776 & 0.641 & 0.003 & 2.781±0.471 & 0.405 \\
\hline
\textbf{PEnet} & 4.0 &3.601±0.564& 0.716 & 0.002 &  $\mathbf{2.081}$±$\mathbf{0.0975}$ & $\mathbf{0.102}$ \\

\textbf{PENN} & 4.0 & \underline{3.800}±\underline{0.495} & \underline{0.575} & 0.002 & 2.121±0.128& 0.144 \\

\textbf{MLP} & 4.0 & 3.692±0.721 & 0.577 & 0.002 & 1.939±0.373 & 0.301 \\
\hline
\textbf{PEnet} & 4.5 & 3.812±0.449 & 0.688& 0.004 & $\mathbf{4.180}$±0.174 & $\mathbf{0.217}$ \\

\textbf{PENN} & 4.5 &\underline{4.009}±\underline{0.426}  & \underline{0.512} & 0.004 & 4.231 ±\underline{0.164} & 0.245 \\

\textbf{MLP} & 4.5 & 3.834±0.652 & 0.684 & 0.004 & 3.591±0.452 & 0.457 \\
\hline
\end{tabular}
\caption*{*The bolded numbers indicate the estimates by the PEnet achieve a more unbiased mean, lower standard deviation (SD), or lower MAE. The numbers are underlined if the previous methods perform better. Table\ref{error2} and \ref{error3} apply a similar setting.}
\end{table*}

To provide further insights into the performance of PEnet, we conducted a comparative analysis with two prominent machine learning-based parameter estimation methods. This evaluation was carried out on a test dataset comprising 5000 samples and involved various combinations of parameters. Specifically, we compared PEnet against the vanilla LSTM-based PENN \cite{wang2022neural}, which has demonstrated comparable performance to traditional LSE \cite{hu2009least} on non-long sequences, as well as an MLP-based method \cite{xie2007estimation}, which is known for its efficiency and accuracy in the context of linear SDEs. The comprehensive results are presented in table \ref{error1}. 
In the five control groups, the proposed PEnet consistently outperformed previous methods in several performance metrics. Particularly, when the value of $\eta$ approached the boundary of the interval, all three methods exhibited significant biases, with PEnet demonstrating the smallest bias. This highlights the superior stability of PEnet compared to other machine learning-based estimation methods, especially in extreme scenarios. Moreover,  PEnet consistently achieved lower mean absolute error (MAE) in estimating the noise intensity $\epsilon$, showcasing its higher competitiveness among machine learning approaches.

\subsection{Case 2 : $\alpha$-stable case}

The $\alpha$ stable-driven SDEs serve as a generalization of the Gaussian-driven SDEs, offering a parameter, $\alpha$, that allows for the adjustment of the noise jump behavior. Consequently, it exhibits greater generality and applicability in practical settings compared to the Gaussian-driven SDEs. In this section, we consider a $\alpha$ stable-driven OU process\cite{uhlenbeck1930theory} $\{X(t)\}_{t\geq 0}$ satisfying the Langevin equation
\begin{align}
\label{alpha stable}
dX(t) = -\eta X(t)dt &+ \epsilon dZ_\alpha(t), \ \ \  t\in[0,+\infty).
\end{align}

\begin{table}[htbp]
    \centering
    \caption{Parameter of $\alpha$ stable driven SDE}
    \begin{tabular}{l|l|l|l}
        \hline
        \textbf{Parameter} & \textbf{Range} & \textbf{Parameter} & \textbf{Range} \\
        \hline
        Spanning Time $T$ & [5, 15] & Length $N$ & [3000, 4000]\\
        Drift $\eta$ & [0, 5] & Noise Intensity $\epsilon$ & [0, 0.05]\\
        Stability index $\alpha$ & [1.01, 2] & & \\
        \hline
    \end{tabular}
    \label{table:params: alpha}
\end{table}

\begin{figure}[htbp]
  \centering
  \begin{minipage}[b]{0.5\textwidth}
    \centering
    \includegraphics[width=0.7\linewidth]{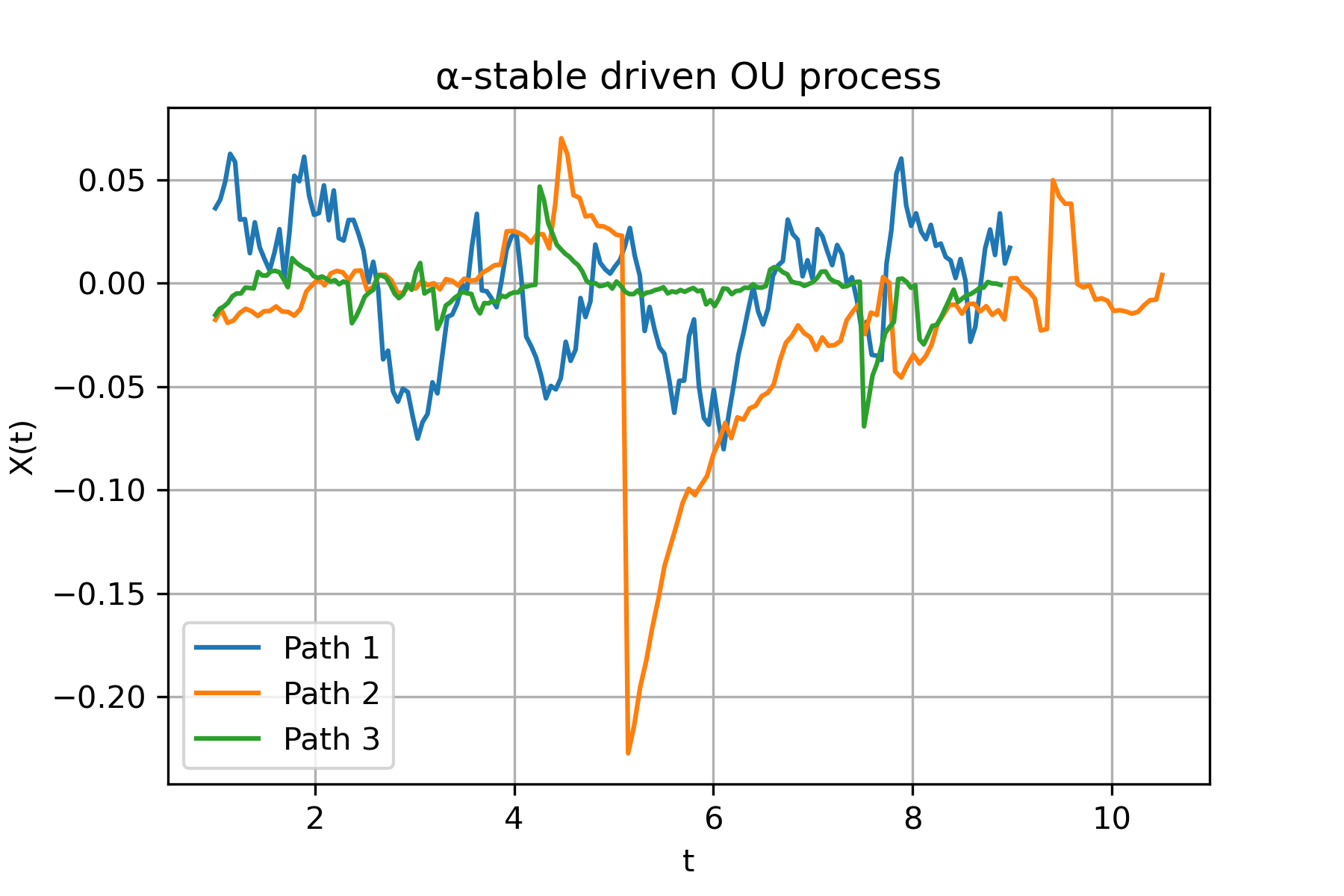}
    \caption{Paths of $\alpha$ stable driven OU processes}
    \label{fig:alpha path}
  \end{minipage}
\end{figure}

\begin{figure}[htbp]
  \centering
  \begin{minipage}[b]{0.5\textwidth}
    \centering
    \includegraphics[width=\linewidth]{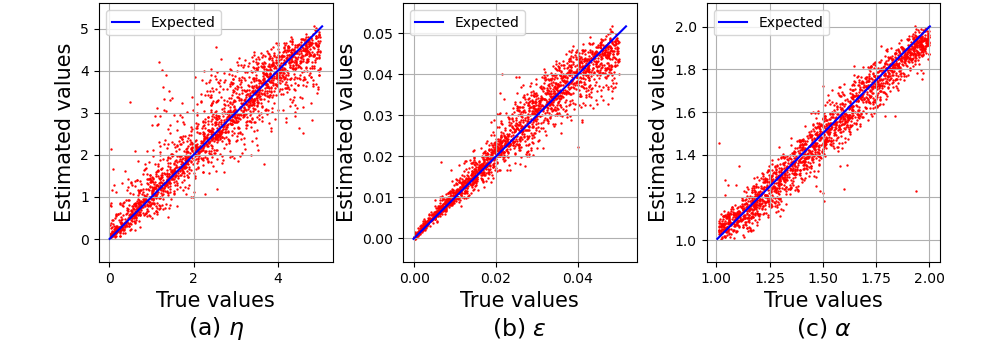}
    \caption{$\hat{\eta}$, $\hat{\epsilon}$ and estimated $\hat{\alpha}$ by the PEnet on 200,000 trajectories of OU process driven by $\alpha$ stable noise.}
    \label{fig:stable predict}
  \end{minipage}
\end{figure}

The parameter set includes three parameters $\mathbf{\Theta} = \{\eta,\epsilon,\alpha\}$ and the output of the network is a three-dimensional vector $\mathbf{\hat{\Theta}} = [\hat{\eta},\hat{\epsilon},\hat{\alpha}]^T$.
Similar to Section \ref{Gaussian exp}, we generated $K_{train} = 200000$ training data samples to train the network. Each sample path was generated using the Euler-Maruyama method with parameters uniformly sampled from the ranges provided in the table \ref{table:params: alpha}. Three paths with different lengths and time spanning are shown in Fig. \ref{fig:alpha path}. 
Compared to Fig. \ref{fig:Gaussian path}, the paths of the $\alpha$-stable noise-driven SDEs display more prominent jumps due to the introduction of the stability index $\alpha$, which controls the jump behaviors of the system. The estimation performance of the trained network is tested on the test set of size $K_{test} = 5000$ and summarized by Fig. \ref{fig:stable predict}.

Compared to Fig. \ref{fig:Gaussian predict}, it is evident that the estimates of noise intensity $\hat{\epsilon}$ exhibit a larger variance compared to the Gaussian case. This can be attributed to the incorporation of the stability index $\alpha$, which adds flexibility to the SDEs model but also introduces increased uncertainty. Similarly to Fig. \ref{fig:Gaussian predict}, estimations of $\eta$ and $\epsilon$ are more accurate within the interior of the interval but exhibit larger bias near the boundaries. Fig. \ref{fig:Gaussian predict} and Fig. \ref{fig:stable predict} collectively indicate that the uncertainty in the estimation $\hat{\eta}$ and $\hat{\epsilon}$ increases linearly with the true values, whereas the estimation of $\alpha$ demonstrates stable variance across the entire interval.    
The performance of PEnet is compared with PENN and MLP. The results are detailed in table \ref{error2}.
\begin{table*}[htbp]
\centering
\caption{Estimation errors of PEnet, PENN and MLP-based method on $\alpha$ stable L\'evy driven OU process}
\label{error2}
\begin{tabular}{|c|c|c|c|c|c|c|c|c|c|c|c|}
\hline
\textbf{Method} & \textbf{$\eta$} & \multicolumn{2}{c|}{\textbf{$\hat{\eta}$}} & \textbf{$\epsilon$} & \multicolumn{2}{c|}{\textbf{$\hat{\epsilon}\times0.01$}}& \textbf{$\alpha$} & \multicolumn{2}{c|}{\textbf{$\hat{\alpha}$}} \\
  & & \textbf{Mean $\pm$ SD} &\textbf{MAE} &  &\textbf{Mean $\pm$ SD}  &\textbf{MAE}&  &\textbf{Mean $\pm$ SD}  &\textbf{MAE} \\
\hline
\textbf{PEnet} & 1.5 & 1.811±0.613 & 0.495 & 0.003 & $\mathbf{3.102}$±$\mathbf{0.314}$ & $\mathbf{0.264}$ & 1.7 & 1.687±$\mathbf{0.0612}$ & $\mathbf{0.0504}$\\

\textbf{PENN} & 1.5 & \underline{1.751}±\underline{0.470} & \underline{0.389} & 0.003 & 3.598±0.452 & 0.381 & 1.7 & \underline{1.691}±0.100 & 0.0812\\

\textbf{MLP} & 1.5 & 1.795±0.740 & 0.576 & 0.003 & 3.297±0.378 & 0.456 & 1.7 & 1.590±0.147 & 0.127\\
\hline
\iffalse
\textbf{PEnet} & 2.5 & 2.768±0.655 & 0.548 & 0.003 & 3.122±$\mathbf{0.316}$ & $\mathbf{0.263}$ & 1.7 & 1.687±$\mathbf{0.0621}$ & $\mathbf{0.0507}$\\

\textbf{PENN} & 2.5 & 2.775±\underline{0.566} & \underline{0.484} & 0.003 & 3.203±0.374 & 0.333 & 1.7 & \underline{1.698}±0.0996 & 0.0801\\

\textbf{MLP} & 2.5 & \underline{2.745}±0.828 & 0.699 & 0.003 & \underline{3.031}±0.608 & 0.499 & 1.7 & 1.581±0.154 & 0.138\\
\hline
\fi
\textbf{PEnet} & 3.5 & 3.570±0.552 & 0.463 & 0.003 & $\mathbf{3.121}$±$\mathbf{0.318}$ & $\mathbf{0.266}$ & 1.7 & 1.684± $\mathbf{0.0620}$ & $\mathbf{0.0513}$\\

\textbf{PENN} & 3.5 & 3.617±\underline{0.510} & \underline{0.432} & 0.003 & 3.123±0.374 & 0.308 & 1.7 & \underline{1.701}±0.101 & 0.0816\\

\textbf{MLP} & 3.5 & \underline{3.504}±0.744 & 0.618 & 0.003 & 2.881±0.600 & 0.493 & 1.7 & 1.571±0.154 & 0.149\\
\hline
\textbf{PEnet} & 2.5 & 2.776±0.655 & 0.549 & 0.002 & $\mathbf{2.081}$±$\mathbf{0.206}$ & $\mathbf{0.206}$ & 1.1 & $\mathbf{1.099}$±$\mathbf{0.0411}$ & $\mathbf{0.032}$\\

\textbf{PENN} & 2.5 & \underline{2.759}±\underline{0.576} & \underline{0.489} & 0.002 & 2.193±0.298 & 0.263 & 1.1 & 1.191±0.105 & 0.109\\

\textbf{MLP} & 2.5 & 2.807±0.825 & 0.721 & 0.002 & 2.142±0.585 & 0.459 & 1.1 & 1.445±0.165 & 0.346\\
\hline
\textbf{PEnet} & 2.0 & $\mathbf{1.997}$±$\mathbf{0.150}$ & $\mathbf{0.104}$ & 0.004 & 4.081±$\mathbf{0.288}$ & $\mathbf{0.242}$ & 1.7 & $\mathbf{1.698}$±$\mathbf{0.0544}$ & $\mathbf{0.0440}$\\

\textbf{PENN} & 2.0 & 2.297±0.339 & 0.337 & 0.004 & \underline{4.064}±0.319 & 0.261 & 1.7 & 1.730±0.0866 & 0.0748\\

\textbf{MLP} & 2.0 & 2.323±0.706 & 0.568 & 0.004 & 3.592±0.517 & 0.486 & 1.7 & 1.551±0.157 & 0.161\\
\hline

\iffalse
\textbf{PEnet} & 2.0 &$\mathbf{2.632}$±0.829 & $\mathbf{0.168}$ & 0.003 & 3.171±$\mathbf{0.439}$ & $\mathbf{0.388}$ & 1.2 & $\mathbf{1.185}$±$\mathbf{0.0503}$ & $\mathbf{0.0431}$\\

\textbf{PENN} & 2.5 & 2.754±\underline{0.358} & 0.327 & 0.003 & \underline{3.144}±0.473 & 0.400 & 1.2 & 1.232±0.0855 & 0.0681\\

\textbf{MLP} & 2.5 & 2.715±0.734 & 0.592 & 0.003 & 3.212±0.687 & 0.605 & 1.2 & 1.419±0.184 & 0.228\\
\hline
\fi

\textbf{PEnet} & 2.5 & $\mathbf{2.666}$±0.488 & $\mathbf{0.380}$ & 0.003 & $\mathbf{3.170}$±$\mathbf{0.412}$ & $\mathbf{0.359}$ & 1.5 & 1.486±$\mathbf{0.0598}$ & $\mathbf{0.0493}$\\

\textbf{PENN} & 2.5 & 2.750± \underline{0.476} & 0.400 & 0.003 & 3.264±0.462 & 0.426 & 1.5 & \underline{1.500}±0.115 & 0.0949\\

\textbf{MLP} & 2.5 & 2.694±0.802 & 0.662 & 0.003 & 3.211±0.639 & 0.537 & 1.5 & 1.519±0.174 & 0.154\\
\hline
\textbf{PEnet} & 2.5 & 2.726±0.684 & 0.571 & 0.001 & $\mathbf{1.032}$±$\mathbf{0.191}$ & $\mathbf{0.147}$ & 1.8 & 1.785±$\mathbf{0.0524}$ & $\mathbf{0.0427}$\\

\textbf{PENN} & 2.5 & 2.754±\underline{0.611} & \underline{0.518} & 0.001 & 1.052±0.231 & 0.177 & 1.8 & \underline{1.788}±0.0875 & 0.0687\\

\textbf{MLP} & 2.5 & \underline{2.720}±0.861 & 0.725 & 0.001 & 1.727±0.894 & 0.834 & 1.8 & 1.611±0.130 & 0.189\\
\hline
\end{tabular}
\end{table*}
Across all six control groups, PEnet consistently outperforms the other two estimation methods in terms of certain metrics. Particularly, when it comes to the random term parameters, PEnet consistently exhibits lower standard deviation and MAE in all control experiments, indicating its superior robustness compared to the other methods. This can be attributed to the effective information compression of jump patterns achieved by the convolutional network in the first stage, which prevents the algorithm from being easily affected by sudden large jumps. It is worth noting that although PENN also achieves relatively small biases in some metrics, especially in the estimation of drift parameters, it trains at a considerably slower pace on long sequence datasets. Taking all of this information into consideration, we can conclude that PEnet is a competitive method compared with other machine learning methods in addressing parameter estimation problems for long sequence SDEs.

\subsection{Case 3 : Student-L\'evy OU process}
\begin{figure}[htbp]
  \centering
  \includegraphics[width=0.3\textwidth]{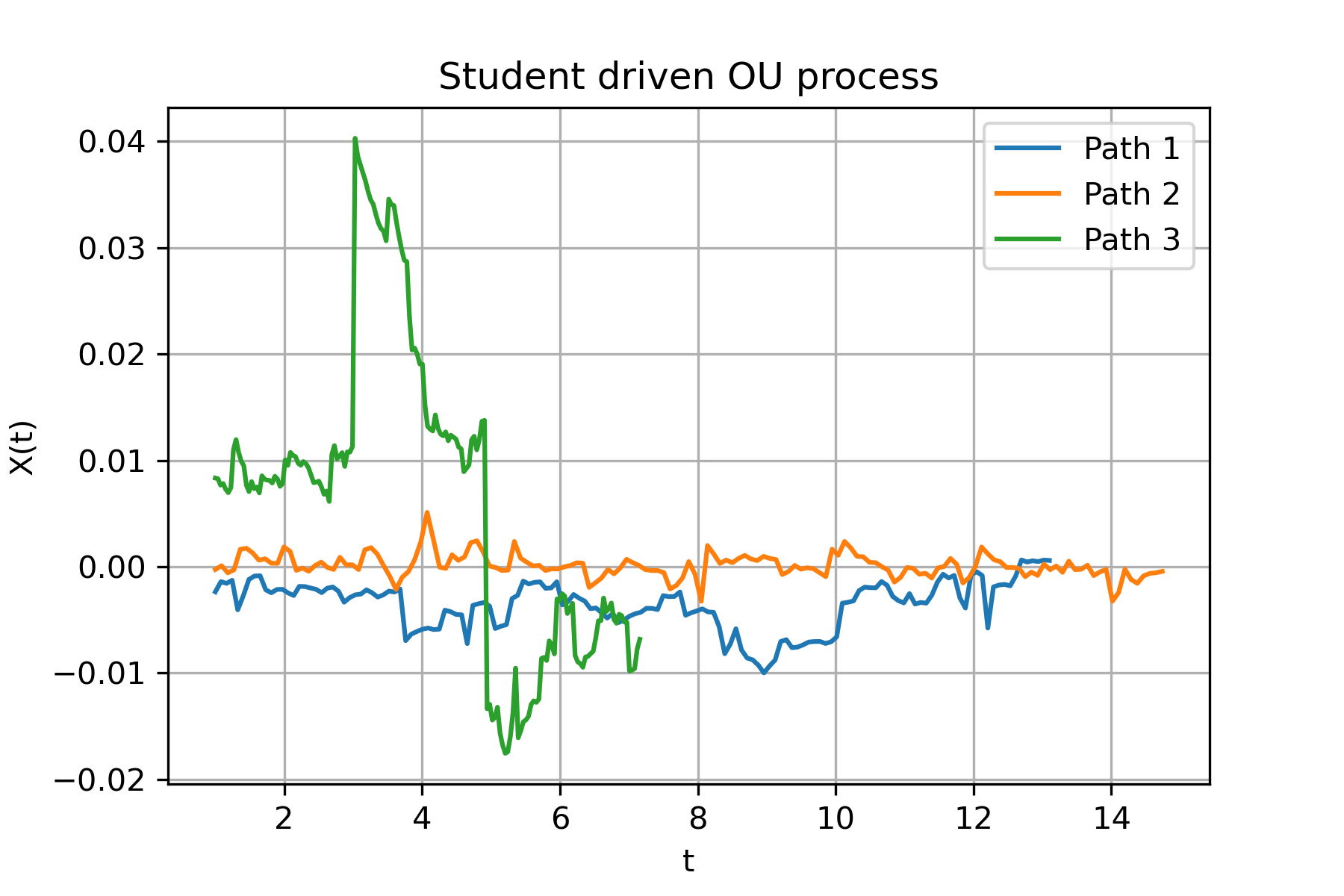}
  \caption{Paths of Student L\'evy driven OU processes}
  \label{fig:Student path}
\end{figure}

\iffalse
Student L\'evy noise, as a generalization of Gaussian noise, provides a better representation of extreme events and outliers. Consequently, it finds wide application in the field of finance, particularly in risk management and financial modeling. However, 
\fi 
The Student L\'evy driven OU process shares remarkable similarities with the $\alpha$ stable case, as evident from Fig. \ref{fig:Student path}. In both scenarios, the system's dynamics are determined by the drift parameter, noise intensity, and the noise parameter controlling jump behavior (stability index $\alpha$ in the case of $\alpha$ stable noise and degrees of freedom $\nu$ for Student-L\'evy noise). This is further confirmed through comparative experiments with other machine learning estimation methods (i.e., PENN and MLP), as the conclusions show no significant differences.

However, a striking contrast lies in the adaptability of the two cases to closed-form statistical methods. Due to the non-closed nature of the convolution in Student-L\'evy noise, its compatibility with traditional closed-form statistical methods differs significantly from $\alpha$ stable case. Consequently, this chapter focuses on the comparison between PEnet and a conventional statistical parameter estimation method, known as Cauchy quasi-maximum likelihood estimation (CQMLE), particularly in estimating the noise parameter $\nu$. 

We consider the following Student L\'evy driven OU process
\begin{align}
\label{alpha stable}
dX(t) = -\eta X(t)dt &+ \epsilon dJ_\nu(t), \ \ \  t\in[0,+\infty),\\
J_\nu(1)&\sim t_\nu(0,1).
\end{align}

\begin{table}[t]
    \centering
    \caption{Parameter of $\alpha$ stable driven SDE}
    \begin{tabular}{l|l|l|l}
        \hline
        \textbf{Parameter} & \textbf{Range} & \textbf{Parameter} & \textbf{Range} \\
        \hline
        Spanning Time $T$ & [3, 15] & Length $N$ & [3000, 4000]\\
        Drift $\eta$ & [0, 5] &  Noise Intensity $\epsilon$ & [0, 0.05]\\
        degrees of freedom $\nu$ & [2.01, 4] && \\
        \hline
    \end{tabular}
    \label{table:params: Student}
\end{table}

The parameter set includes 3 parameters $\mathbf{\Theta} = \{\eta,\epsilon,\nu\}$ and the output of the network is $\mathbf{\hat{\Theta}} = [\hat{\eta},\hat{\epsilon},\hat{\nu}]^T$. 

we generated $K_{train} = 200000$ training data samples to train the network. Each sample path was generated using a modified rejection method \cite{devroye1981computer} with parameters uniformly sampled from the ranges provided in the table \ref{table:params: Student}. Three paths with different lengths and time spanning are shown in Fig. \ref{fig:Student path}. Similar to Fig. \ref{fig:alpha path}, the paths of the Student L\'evy-driven OU process exhibit significant jumps at random times compared to the Gaussian noise case. This characteristic diminishes as the degrees of freedom parameter increases, eventually converging to a smoother path without significant jumps as the degrees of freedom tend to infinity.

\begin{figure}[htbp]
  \centering
  \includegraphics[width=0.45\textwidth]{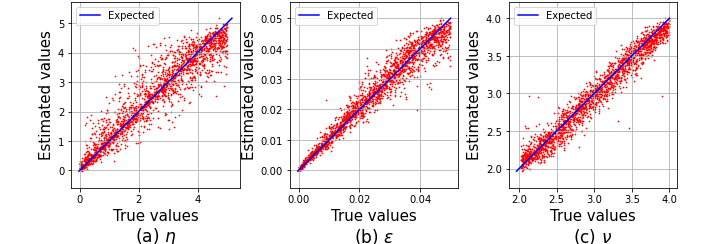}
  \caption{$\hat{\eta}$, $\hat{\epsilon}$ and estimated $\hat{\nu}$ by the PEnet on 200,000 trajectories of OU process driven by Student L\'evy noise.}
  \label{fig:Student predict}
\end{figure}

Based on the findings depicted in Fig. \ref{fig:Student predict},  although the properties of Student L\'evy noise impose substantial obstacles for conventional statistical methodologies, these challenges do not pose a significant impediment to data-driven PEnet. Analogous to the alpha-stable case, $\hat{\eta}$ and $\hat{\epsilon}$ showcase relatively minor bias within the range interior but exhibit pronounced errors near the interval boundaries, accompanied by an increase in uncertainty that scales linearly with the true values. Conversely, the estimation of the degrees of freedom $\nu$, demonstrates a more consistent variance. 

\begin{table}[t]
\centering
\caption{Estimation errors of PEnet, CQMLE on Student L\'evy driven OU process}
\label{error3}
\begin{tabular}{|c|c|c|c|c|c|c|}
\hline
\textbf{Method} & \textbf{$\nu$} & \multicolumn{2}{c|}{\textbf{$\hat{\nu}$}} \\
  & & \textbf{Mean $\pm$ SD} &\textbf{MAE} \\
\hline
\textbf{PEnet} & 2.5 & $\mathbf{2.622}$±$\mathbf{0.0511}$ & $\mathbf{0.0487}$  \\

\textbf{CQMLE} & 2.5 & 2.979±0.616 & 0.642  \\
\hline
\textbf{PEnet} & 3.0 & $\mathbf{3.032}$±$\mathbf{0.0761}$ & $\mathbf{0.0620}$  \\

\textbf{CQMLE} & 3.0 & 3.328±0.818 & 0.765  \\
\hline
\textbf{PEnet} & 3.5 & $\mathbf{3.612}$±$\mathbf{0.0923}$ & $\mathbf{0.0699}$  \\

\textbf{CQMLE} & 3.5 & 3.940±0.944 & 0.891  \\
\hline
\end{tabular}

\end{table}

Table \ref{error3} reveals that the data-driven PEnet outperforms the likelihood-based CQMLE across various metrics. This superiority can be attributed to the fact that many samples in the dataset do not fit the high-frequency observation assumption, leading to significant estimation errors when employing the CQMLE. This result also highlights the substantial advantage of PEnet over statistical approaches in estimating SDEs driven by such complicated noise. 

\section{Conclusion}
SDEs offer powerful tools for modeling random phenomena. However, traditional statistical estimation methods are often limited to Gaussian noise-driven SDEs, highlighting the need for effective data-driven parameter estimation approaches for L\'evy cases. In this study, we propose a CNN-LSTM architecture, named PEnet, that can effectively handle long input sequences with variable lengths. By employing a CNN network to condense the data and extract features through an LSTM network, followed by mapping the features to the parameter space using a fully connected neural network, we achieve more robust and computationally efficient estimation compared to previous methods. Our approach's performance has been validated on SDEs driven by Gaussian, $\alpha$-stable, and Student-L\'evy noise.

As limitations of the PEnet, relatively large estimation errors at the boundaries of the training ranges were observed in the numerical experiments, and the data condensation stage in the CNN may result in information loss, leading to large estimation errors for the drift parameter. To address these limitations, future research will explore the incorporation of model-based knowledge and the adoption of more powerful backbones, such as attention mechanisms, to mitigate these limitations.

 \bibliographystyle{elsarticle-num} 
 \bibliography{lsy_ref}

\begin{thebibliography}{10}
\expandafter\ifx\csname url\endcsname\relax
  \def\url#1{\texttt{#1}}\fi
\expandafter\ifx\csname urlprefix\endcsname\relax\def\urlprefix{URL }\fi
\expandafter\ifx\csname href\endcsname\relax
  \def\href#1#2{#2} \def\path#1{#1}\fi

\bibitem{harris2013flexible}
K.~J. Harris, P.~G. Blackwell, Flexible continuous-time modelling for
  heterogeneous animal movement, Ecological Modelling 255 (2013) 29--37.

\bibitem{gloaguen2018stochastic}
P.~Gloaguen, M.-P. Etienne, S.~Le~Corff, Stochastic differential equation based
  on a multimodal potential to model movement data in ecology, Journal of the
  Royal Statistical Society. Series C (Applied Statistics) 67~(3) (2018)
  599--619.

\bibitem{yilmaz2019stochastic}
A.~Yilmaz, G.~Unal, Stochastic duffing equation in modelling of financial time
  series, International Journal of Dynamics and Control 7 (2019) 1173--1194.

\bibitem{keeling2008methods}
M.~J. Keeling, J.~V. Ross, On methods for studying stochastic disease dynamics,
  Journal of the Royal Society Interface 5~(19) (2008) 171--181.

\bibitem{wang2019continuous}
Y.-S. Wang, P.~G. Blackwell, J.~A. Merkle, J.~R. Potts, Continuous time
  resource selection analysis for moving animals, Methods in Ecology and
  Evolution 10~(10) (2019) 1664--1678.

\bibitem{schneider2014maximum}
G.~Schneider, P.~F. Craigmile, R.~Herbei, Maximum likelihood estimation for
  stochastic differential equations using sequential kriging-based
  optimization, arXiv preprint arXiv:1408.2441 (2014).

\bibitem{delattre2013maximum}
M.~Delattre, V.~GENON-CATALOT, A.~Samson, Maximum likelihood estimation for
  stochastic differential equations with random effects, Scandinavian Journal
  of Statistics 40~(2) (2013) 322--343.

\bibitem{iacus2007sde}
S.~M. Iacus, et~al., sde: Simulation and inference for stochastic differential
  equations (2007).

\bibitem{masuda2019non}
H.~Masuda, Non-gaussian quasi-likelihood estimation of sde driven by locally
  stable l{\'e}vy process, Stochastic Processes and their Applications 129~(3)
  (2019) 1013--1059.

\bibitem{masuda2011quasi}
H.~Masuda, On quasi-likelihood analyses for stochastic differential equations
  with jumps, in: Int. Statistical Inst.: Proc. 58th World Statistical
  Congress, 2011, pp. 83--91.

\bibitem{hu2009least}
Y.~Hu, H.~Long, Least squares estimator for ornstein--uhlenbeck processes
  driven by $\alpha$-stable motions, Stochastic Processes and their
  applications 119~(8) (2009) 2465--2480.

\bibitem{long2009least}
H.~Long, Least squares estimator for discretely observed ornstein--uhlenbeck
  processes with small l{\'e}vy noises, Statistics \& Probability Letters
  79~(19) (2009) 2076--2085.

\bibitem{cheng2020generalized}
Y.~Cheng, Y.~Hu, H.~Long, Generalized moment estimators for $\alpha$-stable
  ornstein--uhlenbeck motions from discrete observations, Statistical Inference
  for Stochastic Processes 23~(1) (2020) 53--81.

\bibitem{fang2022end}
C.~Fang, Y.~Lu, T.~Gao, J.~Duan, An end-to-end deep learning approach for
  extracting stochastic dynamical systems with $\alpha$-stable l{\'e}vy noise,
  Chaos: An Interdisciplinary Journal of Nonlinear Science 32~(6) (2022)
  063112.

\bibitem{xie2007estimation}
Z.~Xie, D.~Kulasiri, S.~Samarasinghe, C.~Rajanayaka, The estimation of
  parameters for stochastic differential equations using neural networks,
  Inverse Problems in Science and Engineering 15~(6) (2007) 629--641.

\bibitem{wang2022neural}
X.~Wang, J.~Feng, Q.~Liu, Y.~Li, Y.~Xu, Neural network-based parameter
  estimation of stochastic differential equations driven by l{\'e}vy noise,
  Physica A: Statistical Mechanics and its Applications 606 (2022) 128146.

\bibitem{kohs2022markov}
L.~K{\"o}hs, B.~Alt, H.~Koeppl, Markov chain monte carlo for continuous-time
  switching dynamical systems, in: International Conference on Machine
  Learning, PMLR, 2022, pp. 11430--11454.

\bibitem{gentili2021characterization}
A.~Gentili, G.~Volpe, Characterization of anomalous diffusion classical
  statistics powered by deep learning (condor), Journal of Physics A:
  Mathematical and Theoretical 54~(31) (2021) 314003.

\bibitem{wagner2017classification}
T.~Wagner, A.~Kroll, C.~R. Haramagatti, H.-G. Lipinski, M.~Wiemann,
  Classification and segmentation of nanoparticle diffusion trajectories in
  cellular micro environments, PloS one 12~(1) (2017) e0170165.

\bibitem{argun2021classification}
A.~Argun, G.~Volpe, S.~Bo, Classification, inference and segmentation of
  anomalous diffusion with recurrent neural networks, Journal of Physics A:
  Mathematical and Theoretical 54~(29) (2021) 294003.

\bibitem{ken1999Levy}
S.~Ken-Iti, L{\'e}vy processes and infinitely divisible distributions,
  Cambridge university press, 1999.

\bibitem{janicki1997approximation}
A.~Janicki, Z.~Michna, A.~Weron, Approximation of stochastic differential
  equations driven by $\alpha$-stable l{\'e}vy motion, Applicationes
  Mathematicae 24~(2) (1997) 149--168.

\bibitem{applebaum2009levy}
D.~Applebaum, L{\'e}vy processes and stochastic calculus, Cambridge university
  press, 2009.

\bibitem{massing2019stochastic}
T.~P.~G. Massing, Stochastic properties of student-l{\'e}vy processes with
  applications, Ph.D. thesis, Universit{\"a}t Duisburg-Essen (2019).

\bibitem{massing2018simulation}
T.~Massing, Simulation of student--l{\'e}vy processes using series
  representations, Computational Statistics 33~(4) (2018) 1649--1685.

\bibitem{tang2020rethinking}
W.~Tang, G.~Long, L.~Liu, T.~Zhou, J.~Jiang, M.~Blumenstein, Rethinking 1d-cnn
  for time series classification: A stronger baseline, arXiv preprint
  arXiv:2002.10061 (2020) 1--7.

\bibitem{li2021survey}
Z.~Li, F.~Liu, W.~Yang, S.~Peng, J.~Zhou, A survey of convolutional neural
  networks: analysis, applications, and prospects, IEEE transactions on neural
  networks and learning systems (2021).

\bibitem{hochreiter1997long}
S.~Hochreiter, J.~Schmidhuber, Long short-term memory, Neural computation 9~(8)
  (1997) 1735--1780.

\bibitem{greff2016lstm}
K.~Greff, R.~K. Srivastava, J.~Koutn{\'\i}k, B.~R. Steunebrink, J.~Schmidhuber,
  Lstm: A search space odyssey, IEEE transactions on neural networks and
  learning systems 28~(10) (2016) 2222--2232.

\bibitem{yu2019review}
Y.~Yu, X.~Si, C.~Hu, J.~Zhang, A review of recurrent neural networks: Lstm
  cells and network architectures, Neural computation 31~(7) (2019) 1235--1270.

\bibitem{clevert2015fast}
D.-A. Clevert, T.~Unterthiner, S.~Hochreiter, Fast and accurate deep network
  learning by exponential linear units (elus), arXiv preprint arXiv:1511.07289
  (2015).

\bibitem{ioffe2015batch}
S.~Ioffe, C.~Szegedy, Batch normalization: Accelerating deep network training
  by reducing internal covariate shift, in: International conference on machine
  learning, pmlr, 2015, pp. 448--456.

\bibitem{nair2010rectified}
V.~Nair, G.~E. Hinton, Rectified linear units improve restricted boltzmann
  machines, in: Proceedings of the 27th international conference on machine
  learning (ICML-10), 2010, pp. 807--814.

\bibitem{kloeden1992stochastic}
P.~E. Kloeden, E.~Platen, P.~E. Kloeden, E.~Platen, Stochastic differential
  equations, Springer, 1992.

\bibitem{weron1996chambers}
R.~Weron, On the chambers-mallows-stuck method for simulating skewed stable
  random variables, Statistics \& probability letters 28~(2) (1996) 165--171.

\bibitem{chambers1976method}
J.~M. Chambers, C.~L. Mallows, B.~Stuck, A method for simulating stable random
  variables, Journal of the american statistical association 71~(354) (1976)
  340--344.

\bibitem{devroye1981computer}
L.~Devroye, On the computer generation of random variables with a given
  characteristic function, Computers \& Mathematics with Applications 7~(6)
  (1981) 547--552.

\bibitem{paszke2019pytorch}
A.~Paszke, S.~Gross, F.~Massa, A.~Lerer, J.~Bradbury, G.~Chanan, T.~Killeen,
  Z.~Lin, N.~Gimelshein, L.~Antiga, et~al., Pytorch: An imperative style,
  high-performance deep learning library, Advances in neural information
  processing systems 32 (2019).

\bibitem{kingma2014adam}
D.~P. Kingma, J.~Ba, Adam: A method for stochastic optimization, arXiv preprint
  arXiv:1412.6980 (2014).

\bibitem{uhlenbeck1930theory}
G.~E. Uhlenbeck, L.~S. Ornstein, On the theory of the brownian motion, Physical
  review 36~(5) (1930) 823.

\end{thebibliography}

\end{document}